\documentclass{article}

\usepackage[utf8]{inputenc}
\usepackage[english]{babel}
\usepackage{graphicx} 
\usepackage{subcaption}
\usepackage{amsmath} 
\usepackage{amssymb}  
\usepackage{algorithm}
\usepackage{algorithmic} 

\usepackage{hyperref}

\newtheorem{theorem}{Example}

\newcommand{\mbs}[1]{\boldsymbol{#1}}
\newcommand{\mbf}[1]{\mathbf{#1}}
\newcommand{\parm}{\mbs{\theta}}
\newcommand{\parmset}{\Theta}
\newcommand{\prbsc}{\pi}
\newcommand{\prb}{\mbs{\prbsc}}
\newcommand{\prbset}{\Pi}
\newcommand{\truepdf}{p_{o}}
\newcommand{\corrpdf}{q}
\newcommand{\obspdf}{p}
\newcommand{\corr}{\epsilon}
\newcommand{\corrub}{\widetilde{\corr}}
\newcommand{\z}{\mbs{z}}
\newcommand{\data}{\mathcal{D}}
\newcommand{\bestmodel}{\parm^{\star}}
\newcommand{\plgest}{\widehat{\parm}}
\newcommand{\R}{\mathbb{R}}

\newcommand{\loss}{\ell}
\newcommand{\erm}{\textsc{Erm}}
\newcommand{\risk}{R}
\newcommand{\entr}{\mathbb{H}}

\newcommand{\optparm}{\widehat{\mbs{\parm}}}
\newcommand{\optprb}{\widehat{\mbs{\prb}}}
\newcommand{\y}{y}
\newcommand{\x}{x}
\newcommand{\xvec}{\mbs{\x}}
\newcommand{\rrm}{\textsc{Rrm}}
\newcommand{\tpdf}{t}

\newcommand{\severalg}{\textsc{Sever}}
\newcommand{\kkt}{\textsc{KKT}}

\DeclareMathOperator{\E}{\mathbb{E}}
\DeclareMathOperator*{\argmin}{arg\,min}
\DeclareMathOperator*{\argmax}{arg\,max}
\DeclareMathOperator{\T}{\top}

\usepackage[accepted]{icml2020}

\icmltitlerunning{Robust Risk Minimization for Statistical Learning}

\begin{document}

\twocolumn[
\icmltitle{Robust Risk Minimization for Statistical Learning}



\icmlsetsymbol{equal}{*}

\begin{icmlauthorlist}
\icmlauthor{Muhammad Osama}{uu}
\icmlauthor{Dave Zachariah}{uu}
\icmlauthor{Peter Stoica}{uu}
\end{icmlauthorlist}

\icmlaffiliation{uu}{Department of Information Technology, Uppsala University, Sweden}

\icmlcorrespondingauthor{Muhammad Osama}{muhammad.osama@it.uu.se}
\icmlcorrespondingauthor{Dave Zachariah}{dave.zachariah@it.uu.se}

\icmlkeywords{Machine Learning, ICML}

\vskip 0.3in
]



\printAffiliationsAndNotice{\icmlEqualContribution} 

\begin{abstract}
We consider a general statistical learning problem where an unknown fraction of the training data is corrupted. We develop a robust learning method that only requires specifying an upper bound on the corrupted data fraction. The method minimizes a risk function defined by a non-parametric distribution with unknown probability weights. We derive and analyse the optimal weights and show how they provide robustness against corrupted data. Furthermore, we give a computationally efficient coordinate descent algorithm to solve the risk minimization problem. We demonstrate the wide range applicability of the method, including regression, classification, unsupervised learning and classic parameter estimation, with state-of-the-art performance. 
\end{abstract}


\begin{figure*}[h!]
    \centering
    \begin{subfigure}{0.45\linewidth}
    \includegraphics[width=0.9\linewidth]{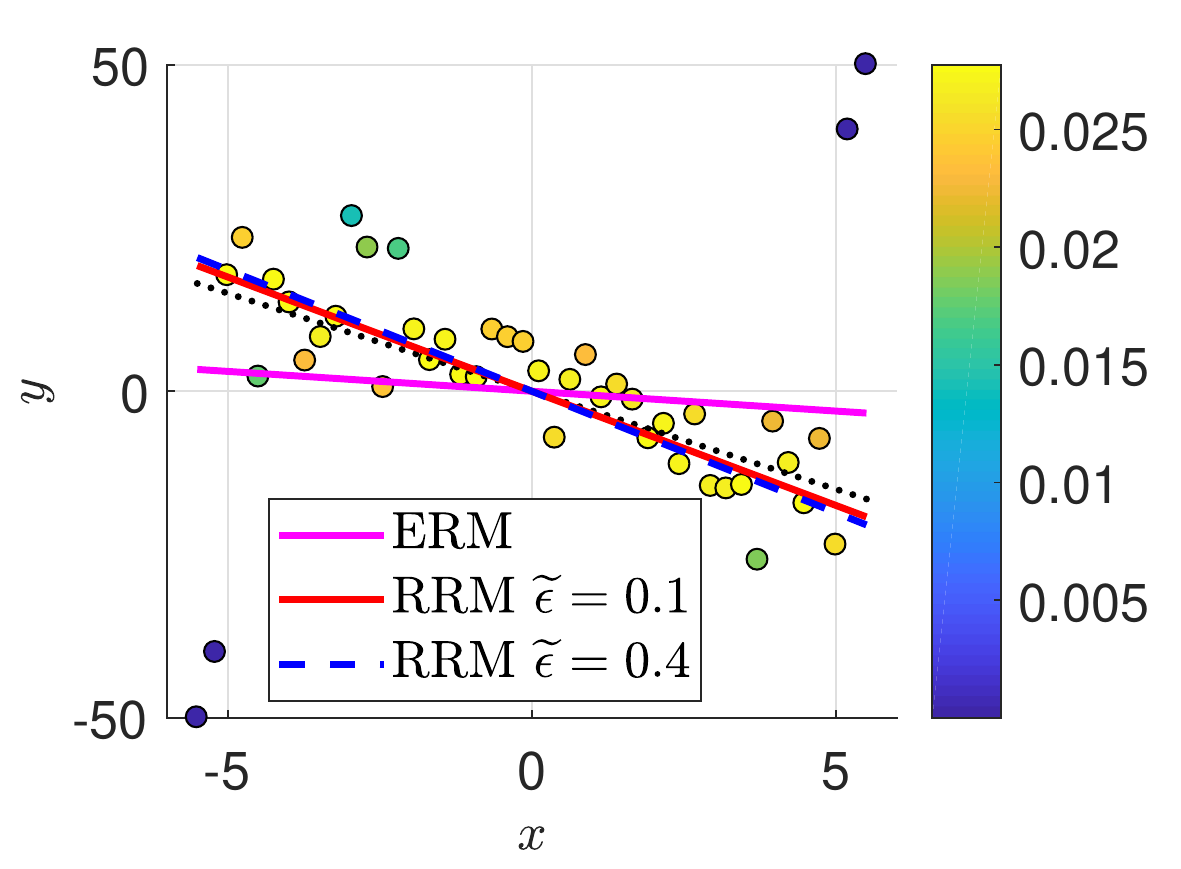}    \caption{}
    \label{fig:probillus_linReg}
    \end{subfigure}
    \begin{subfigure}{0.45\linewidth}
    \includegraphics[width=0.9\linewidth]{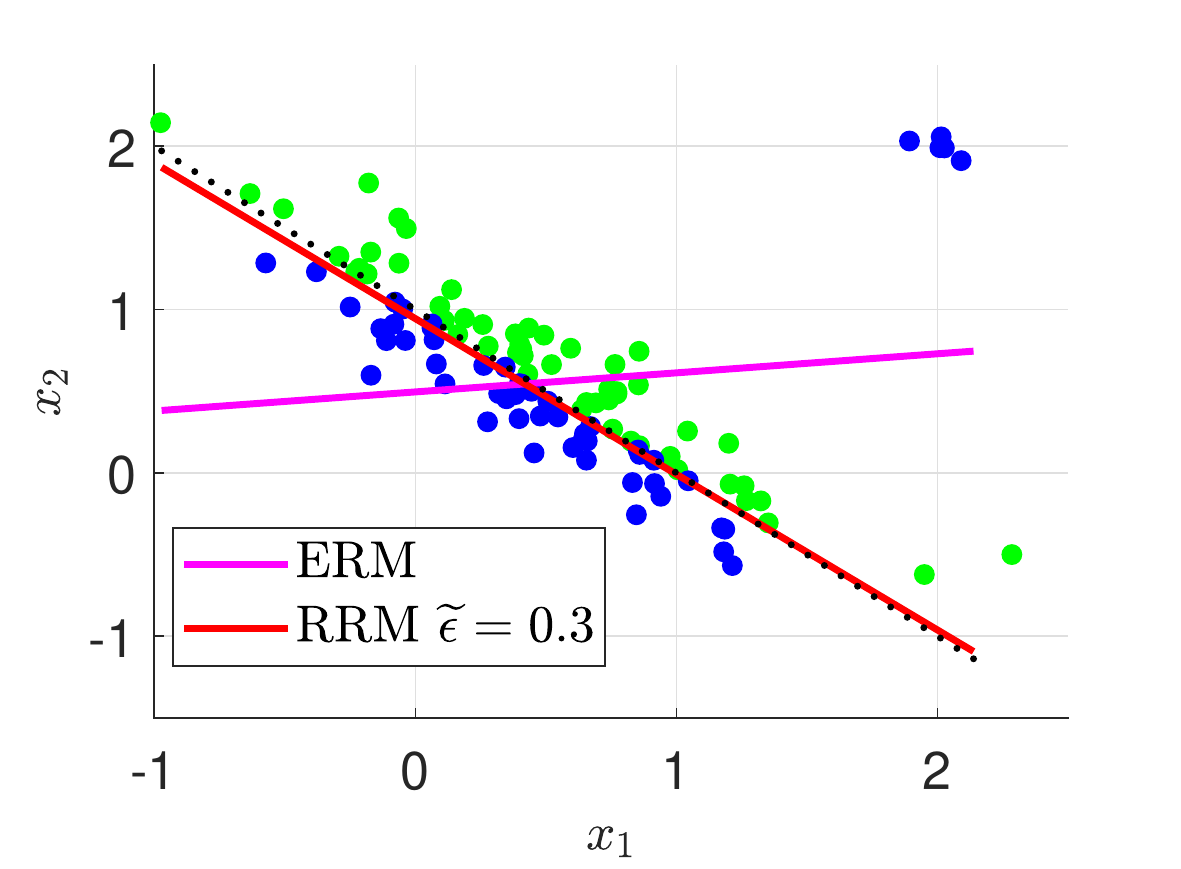}
    \caption{}
    \label{fig:probillus_logre}
    \end{subfigure}
    \caption{Illustration of statistical learning from data with an unknown fraction $\corr$ of corrupted samples.
    (a) Regression, where $\parm$ parameterizes a linear model. Learned models using standard least squares (\erm{}) vs. proposed method (\rrm{}) using two different upper bounds $\corrub$ on $\corr$. The target $\bestmodel$ is illustrated in black dots. The proposed method learns and assigns weights to each observed data point (shown in color scale). The weights of the outlier points are very small and, in turn, contribute marginally to learning the regression line. (b) Binary classification, where $\parm$ parameterizes a separating hyperplane. The label data $y$ is blue for `0' and green for '1'.
    Learned models using standard logistic regression (\erm{})  vs.  proposed method (\rrm{}) using an upper bound $\corrub$. The target $\bestmodel$ is illustrated in black dots.}
    \label{fig:probillus}
\end{figure*}

\section{Introduction}

Statistical learning problems encompass regression, classification, unsupervised learning and parameter estimation \cite{bishop2006pattern}. The common goal is to find a model, indexed by a parameter $\parm$, that minimizes some loss function $\loss_{\parm}(\z)$ on average, using training data $\data=\{\z_1, \dots, \z_n \}$. The loss function is chosen to target data from a class of distributions, denoted $\mathcal{P}_o$.

It is commonly assumed that the training data is drawn from some distribution $\truepdf(\z) \in \mathcal{P}_o$. In practice, however, training data is often corrupted by outliers, systematic mislabeling, or even an adversary. Under such conditions, standard learning methods degrade rapidly \cite{athalye18adversial, goodfellow2015, gu2017badnets, zoubir2018robust}. See Figure~\ref{fig:probillus} for an illustration. Here we consider the Huber contamination model which is capable of modeling the inherent corruption of data and is common in the robust statistics literature \cite{huber1992robust, huber2011robust, maronna2019robust}. Specifically, the training data is assumed to be drawn from the unknown mixture distribution
\begin{equation}\label{eq:obspdf}
\boxed{\obspdf(\z)=(1-\corr)\truepdf(\z)+\corr\corrpdf(\z),}
\end{equation}
so that roughly $\corr n$ samples come from a corrupting distribution $\corrpdf(\z) \not  \in \mathcal{P}_o$. The fraction of outliers, $\corr$, may range between $1-10\%$ in routine datasets, but in data collected with less dedicated effort or under time constraints $\corr$ can easily exceed 10\% \citep[ch.~1]{hampel2011robust}.

In the robust statistics literature,  several methods have been developed for various applications. A classical approach is to modify a given loss function $\loss_{\parm}(\z)$ so as to be less sensitive to outliers  \cite{huber2011robust, maronna2019robust, zoubir2018robust}. Some examples of such  functions are the Huber and Tukey  loss functions \cite{huber1992robust,tukey1977}. Another approach is to try and identify the corrupted points in the training data based on some criteria and then remove them \cite{klivans2009learning,bhatia2015robust,bhatia2017consistent,awasthi2017linSep,paudice2018detection}. For example, for mean and covariance estimation of $\z \sim \truepdf(\z)$, the method presented in \cite{diakonikolas2017robust} identifies corrupted points by projecting the training data onto an estimated dominant signal subspace and then compares the magnitude of the projected data against some threshold. 
The main limitation of the above approaches is that they are problem-specific and must be tailored to each learning problem. In addition, even for fairly simple learning problems, such as inferring the mean, the robust estimators can be computationally demanding \cite{diakonikolas2019recent,bernholt2006robust,hardt2013algorithms}.

Recent work has been directed toward developing more general and tractable methods for robust statistical learning that is applicable to a wide range of loss functions \cite{diakonikolas19sever, prasad2018robust, charikar2017untrusted}. These state-of-the-art methods do, however, exhibit some important limitations relating to the choice of certain tuning parameters. For instance, the two-step method in \cite{charikar2017untrusted} uses a regularization parameter which depends on the unknown $\corr$ that the user may not be able to  specify precisely. Similarly, under certain conditions, there exists a parameter setting for the method in \cite{diakonikolas19sever} that  yields performance guarantees, but there is no practical means or criterion for how to tune this parameter. The cited methods rely on removing data points based on some score function and comparing to a specified threshold. The extent to which the choice of scoring function is problem dependent is unknown and the choice of threshold depends in practice on some user-defined parameter. 


The main contribution of this paper is a general robust method with the following properties:
\begin{itemize}
    \item it is applicable to any statistical learning problem that minimizes an expected loss function,
    \item it requires only specifying an upper bound on the corrupted data fraction $\corr$, 
    \item it is formulated as a minimization problem that can be solved efficiently using a blockwise algorithm.
\end{itemize}
We illustrate and evaluate the robust method in several standard statistical learning problems.

\section{Problem formulation}

Consider a set of models indexed by a parameter $\parm \in \parmset$. The predictive loss of a model $\parm$ is denoted $\loss_{\parm}(\z)$, where $\z \sim \truepdf(\z)$ is a randomly drawn datapoint. The target model is that which minimizes the expected loss, or \emph{risk}, i.e.,
\begin{equation}\label{eq:bestmodel}
    \bestmodel=\argmin_{\parm \in \parmset}~\E_{\truepdf}\big[\loss_{\parm}(\z)\big],
\end{equation}
Because the target distribution $\truepdf(\z)$ is typically unknown, we use $n$ independent samples $\data~=~\{\z_i\}_{i=1}^n$ drawn from $\obspdf(\z)$ in \eqref{eq:obspdf}. 
A common learning strategy is to find the empirical risk minimizing (\erm) parameter vector
\begin{equation}\label{eq:emp risk min}
    \plgest_{\textsc{erm}}=\argmin_{\parm \in \parmset}~\frac{1}{n}\sum_{i=1}^{n}\loss_{\parm}(\z_i)
\end{equation}

\begin{theorem}
In regression problems, data consists of features and outcomes, $\z = (\xvec, \y)$, and $\parm$ parameterizes a predictor $\widehat{\y}_{\parm}(\xvec)$. The standard loss function $\loss_{\parm}(\z) = (\y - \widehat{\y}_{\parm}(\xvec))^2$ targets distributions with thin-tailed noise.
\end{theorem}

\begin{theorem}
In general parameter estimation problems, a standard loss function is $\loss_{\parm}(\z) = - \ln p_{\parm}(\z)$, which targets distributions spanned by $\obspdf_{\parm}(\z)$. For this choice of loss function, \eqref{eq:emp risk min} corresponds to the maximum likelihood estimator.
\end{theorem}

In real applications, a certain fraction $\corr \in [0,1)$ of the  data is \emph{corrupted} such that the risk under the unknown data-generating process $\obspdf(\z)$ exceeds the minimum in \eqref{eq:bestmodel}. That is,
\begin{equation}
\E_{\obspdf}[\loss_{\parm}(\z)] \; \geq \; \E_{\truepdf}[\loss_{\bestmodel}(\z)], \quad \forall \parm,
\label{eq:obspdf_risk}
\end{equation}
with equality if and only if $\corr = 0$. 
Under such corrupted data conditions, \erm{} degrades rapidly as $\obspdf(\z)$ diverges from $\truepdf(\z)$. While $\corr$ in \eqref{eq:obspdf} is unknown, it can typically be upper bounded, that is, $\corr\leq\corrub$ \citep[ch.~1]{hampel2011robust}. This implies that there are effectively at least  $(1-\corrub)n$ uncorrupted samples in $\data$. For learning problems with unregularized loss functions, this sample size must therefore be at least as great as the dimension of $\parm$. Our goal is to formulate a general method of risk minimization, which given $\data$ and $\corrub$, learns a model $\parm$ that is robust against corrupted training samples.

\begin{figure}[t!]
    \centering
    \begin{subfigure}{0.9\linewidth}
    \includegraphics[width=0.9\linewidth]{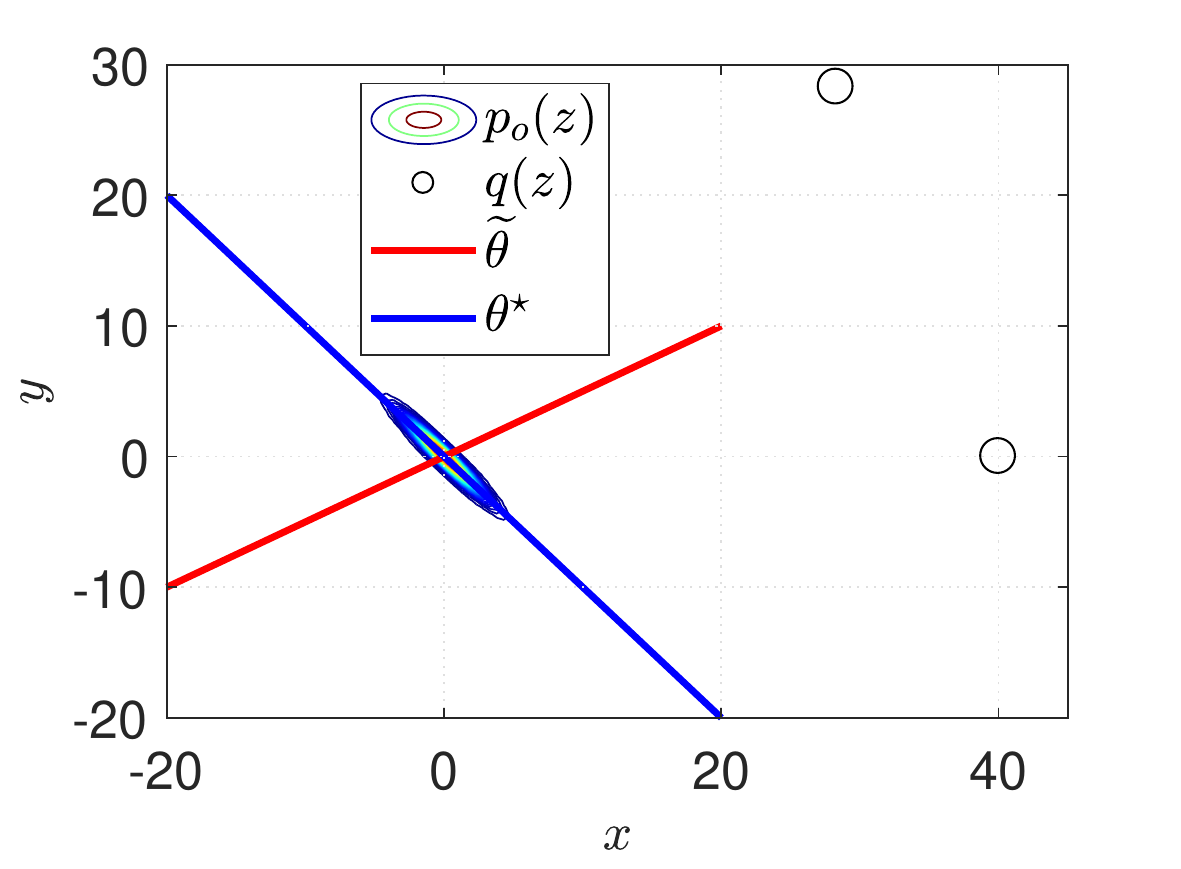}
    \caption{}
    \label{fig:leverage 1}
    \end{subfigure}
    \begin{subfigure}{0.9\linewidth}
    \includegraphics[width=0.9\linewidth]{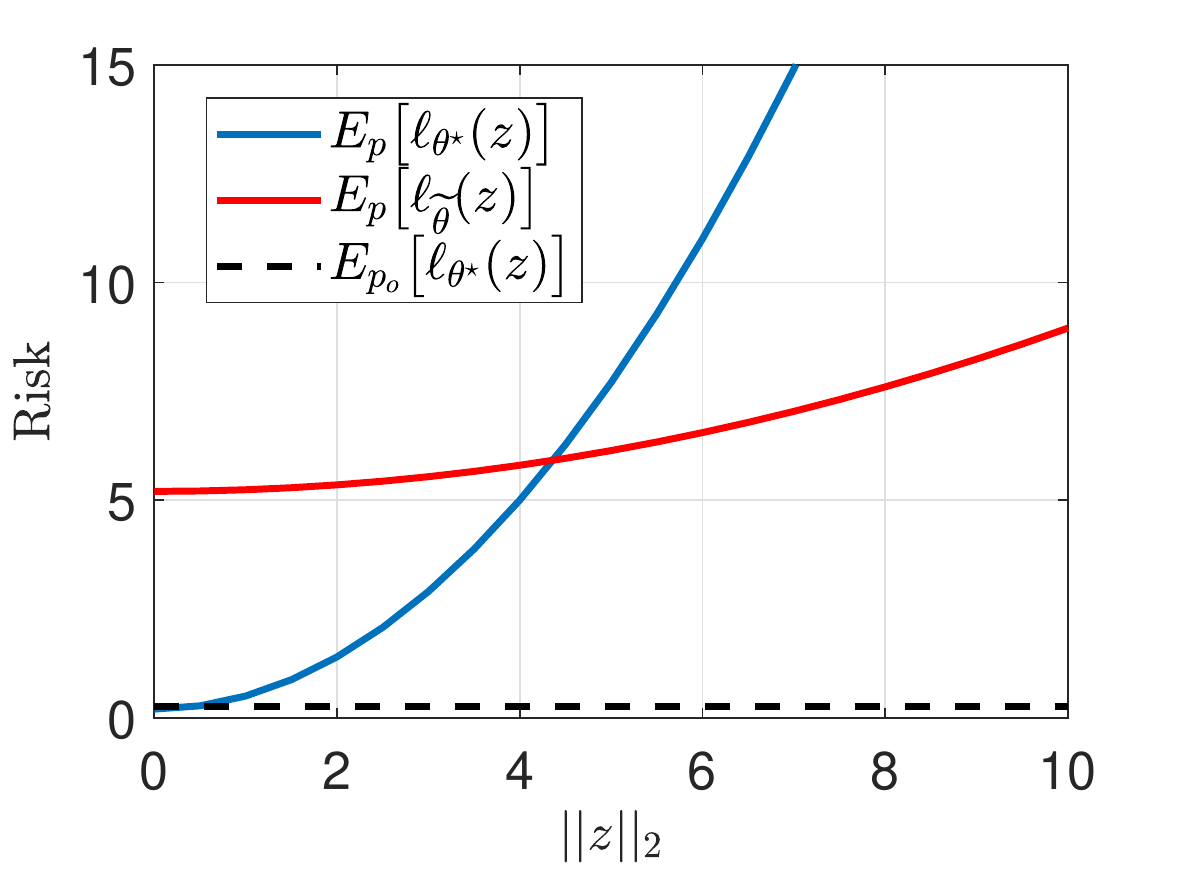}
    \caption{}
    \label{fig:leverage risk}
    \end{subfigure}
    \caption{}
    \label{fig:leverage}
\end{figure}

\begin{theorem}
To illustrate the degradation of \erm{} as $\obspdf(\z)$ diverges from $\truepdf(\z)$, consider a linear regression problem where $\z=(x, y) \sim \obspdf(\z)$, using the squared-error loss. Fig.~\ref{fig:leverage 1} illustrates the target distribution $\truepdf(\z)$, which is a zero-mean two-dimensional Gaussian, and $\corrpdf(\z)$ as a point mass distribution generating corrupted leverage points at equal distances to the origin. The figure also illustrates two contrasting regression models $\{ \widetilde{\parm}, \bestmodel \}$. Fig.~\ref{fig:leverage risk} shows how the risk $\E_p[\loss_{\parm}(\z)]$ increases for these two models, as the distance of the corrupted points from the training data increases. In the large-sample case, \erm{} minimizes this risk and will therefore drastically degrade as it opts for $\widetilde{\parm}$ over $\bestmodel$ at a certain distance. We note, however, that $\E_{\obspdf}\big[\ell_{\widetilde{\parm}}(\z)\big]\geq\E_{\truepdf}\big[\ell_{\bestmodel}(\z)\big]$ (more generally, \eqref{eq:obspdf_risk} holds). That is, by retaining only an $1-\corr$ fraction of uncorrupted data from $\truepdf(\z)$, it is possible to identify $\bestmodel$ with lower risk than an alternative model $\parm$. This principle will be exploited in the next section.      
\end{theorem}
\section{Method}
We define the risk evaluated at a distribution $\obspdf_{\prb}(\z)$ as
\begin{equation}\label{eq:propose risk}
    \risk(\parm, \prb)=\E_{\obspdf_{\prb}}\big[\loss_{\parm}(\z)\big]
\end{equation}
We consider the following nonparametric class of distributions, indexed by $\prb$,
\begin{equation}\label{eq:proposed obspdf model}
\left\{  \obspdf_{\prb}\: : \: \obspdf_{\prb}(\z)=\sum_{i=1}^{n}\prbsc_i\delta(\z-\z_i), \: \prb \in \prbset \right\}
\end{equation}
where  $\z_i \in \data$ and the weights belong to the simplex $\prbset= \left\{ \prb \in \mathbb{R}^n_+ : \mbf{1}^{\T}\prb~=~1 \right\}$. Let the entropy of $\obspdf_{\prb}(\z)$ be denoted as
$$\entr(\prb) \triangleq -\sum_i \prbsc_i \ln \prbsc_i \; \in \;  [0, \: \ln n],$$
then it is readily seen that \erm{} in \eqref{eq:emp risk min} minimizes the risk $\risk(\parm, \prb)$ under the maximum-entropy distribution $\prb = n^{-1}\mbf{1}$ \cite{cover2012elements}. When $\corr =0$, it is well known this choice yields an asymptotically  consistent estimate of $\bestmodel$ under standard regularity conditions.

\subsection{Robust risk minimization} 

If the support of $\obspdf_{\prb}$ only covers $(1-\corr) n$ samples drawn from $\truepdf$, its maximum entropy is at least $\ln[ (1 - \corrub)n]$ since $\corrub \geq \corr$. The risk $\risk(\bestmodel, \prb)$ will then tend  to be lower than  $\risk(\bestmodel, n^{-1}\mbf{1})$, since the latter includes $\corr n$ corrupted samples, cf. \eqref{eq:obspdf_risk}. We propose learning $\parm$ by utilizing the distribution in \eqref{eq:proposed obspdf model} that yields the minimum risk, subject to its entropy $\entr(\prb)$ being at least $\ln[ (1 - \corrub)n]$. That is, the following robust risk minimization (\rrm{}) approach
\begin{equation}\label{eq:main optimization problem}
\boxed{\plgest_{\textsc{rrm}}  = \argmin_{\parm \in \parmset} \:  \min_{\prb \in \prbset \: : \: \entr(\prb)\geq\ln\left[(~1-\corrub~)n\right]}~\risk(\parm, \prb) }
\end{equation}
The entropy constraint ensures that $\parm$ is learned using an effective sample size of $(1 - \corrub)n$. 

We now study the optimal weights $\prb^{\star}(\parm)$ of the inner problem in \eqref{eq:main optimization problem} to understand how the proposed method yields robustness against corrupted samples. Note that the objective function in \eqref{eq:main optimization problem}, $\risk(\parm, \prb)= \sum^n_{i=1} \pi_i \loss_{\parm}(\z_i)$, is linear in $\prb$, and that the entropy constraint is convex. The inner optimization problem satisfies Slater's condition, since $\prb~=~n^{-1}\mbf{1}$ is a strictly feasible point. Hence strong duality holds and the optimal weights $\prb^{\star}$ can be obtained using the Karush-Kuhn-Tucker (\kkt) conditions \cite{boyd2004convex}. The Lagrangian of the optimization problem with respect to $\prb$ equals
\begin{equation*}
\begin{split}
    L(\prb,\lambda,\nu)=&\sum_{i=1}^{n}\prbsc_{i}\loss_{\parm}(\z_{i})~+~ \lambda(\ln\big[(~1-\corrub~)n\big]-\entr(\prb))\\
                        &+~\nu(\mbf{1}^{\top}\prb-1)
\end{split}
\end{equation*}
where $\lambda$ and $\nu$ are the dual variables. The \kkt{} conditions can then be expressed as 
\begin{equation}
    \begin{split}
        \nabla_{\prb}~L(\prb^{\star},\lambda^{\star},\nu^{\star})=~0,\quad\mbf{1}^{\top}\prb^{\star}=~1, \quad\lambda^{\star}\geq~0,\\
        \lambda^{\star}(\ln\big[(~1-\corrub~)n\big]-\entr(\prb^{\star}))=~0,\\
        \entr(\prb^{\star})\geq~\ln\big[(~1-\corrub~)n\big]\\
    \end{split}
\end{equation}
Here $(\lambda^{\star},\nu^{\star})$ are the dual optimal solutions. Solving the above \kkt{} conditions leads to the following optimal weights,
\begin{equation} \label{eq: opt weights fixed theta}
    \prbsc_{i}^{\star}(\parm)=c^{\star}\exp\left[-\frac{\loss_{\parm}(\z_{i})}{\lambda^{\star}(\corrub)} \right] \geq 0,  \; i = 1, \dots, n,
\end{equation}
where $c^{\star}$ is a proportionality constant which ensures that the probability weights sum to $1$. It is immediately seen that $\prb^\star(\parm)$ downweights the data points with high losses for any given $\parm$. As the corruption fraction bound $\corrub$ vanishes, the attenuation factor $1/\lambda^{\star}(\corrub)$ goes to zero, thereby yielding uniform weights $\prb^\star(\parm) \rightarrow n^{-1}\mbf{1}$ as expected.

Plugging $\prb^\star(\parm)$ back into \eqref{eq:main optimization problem} yields the equivalent concentrated problem
\begin{equation}\label{eq:concen obj}
    \min_{\parm \in \parmset}~~\sum_{i=1}^{n} \exp\left[-\frac{\loss_{\parm}(\z_{i})}{\lambda^{\star}(\corrub)}\right]\loss_{\parm}(\z_{i}),
\end{equation}
which focuses the learning of $\parm$ on the set of $(1-\corrub)n$ samples with lowest losses. 
By downweighting corrupted samples that increase the risk at every $\parm$, the equivalent problem \eqref{eq:main optimization problem} provides robustness against outliers in $\data$ in the learning of $\bestmodel$ (see  \eqref{eq:obspdf_risk}). This is achieved without tailoring a new robustified loss function or tuning a loss-specific user parameter to a given problem. Instead, the user needs only specify an upper bound of the corruption fraction, $\corrub$.

\subsection{Blockwise minimization algorithm} 
We now propose an efficient computational method of finding a solution of \eqref{eq:main optimization problem}.
Given fixed parameters $\widetilde{\parm}$ and $\widetilde{\prb}$, we define for given $\widetilde{\parm}$
\begin{equation}\label{eq:sub problem 1}
\optprb(\widetilde{\parm})=
        \underset{\prb \in \prbset \: : \: \entr(\prb)\geq\ln\left[(~1-\corrub~)n\right]}{\argmin}~\risk(\widetilde{\parm}, \prb),
\end{equation}
which is the solution to a convex optimization problem and can be computed efficiently using standard numerical packages, e.g. barrier methods \cite{grant2014cvx} which have polynomial-time complexity. For a given $\widetilde{\prb}$, the minimizer
\begin{equation}\label{eq:sub problem 2}
\optparm(\widetilde{\prb})=\argmin_{\parm \in \parmset}~\risk(\parm, \widetilde{\prb}),
\end{equation}
is the solution to a standard weighted risk minimization problem. Solving both problems in a cyclic manner constitutes blockwise coordinate descent method 
which we summarize in Algorithm~\ref{algo:RRM}. When the parameter set $\parmset$ is closed and convex, the algorithm is guaranteed to converge to a critical point of \eqref{eq:main optimization problem}, see \cite{grippo2000convergence}.
\begin{algorithm}
\caption{Robust Risk Minimization (\rrm)} \label{algo:RRM}
\begin{algorithmic}[1]
\STATE Input: $\data$ and $\corrub$
\STATE Set $k~:=~0$  and $\prb^{(0)}~=~n^{-1}\mbf{1}$
\STATE \textbf{repeat}
\STATE $\parm^{(k+1)}~=~\optparm(\prb^{(k)})$
\STATE $\prb^{(k+1)}~=~\optprb(\parm^{(k+1)})$
\STATE $k~:=~k~+~1$
\STATE \textbf{until convergence}
\STATE Output: $\optparm~=~\parm^{(k)}$,~$\optprb~=~\prb^{(k)}$
\end{algorithmic}
\end{algorithm}\\
The general form of the proposed method renders it applicable to a diverse range of learning problems in which $\erm$ is conventionally used. In the next section, we illustrate the performance and generality of the proposed method using numerical experiments for different supervised and unsupervised machine learning problems. The code for the different experiments can be found at \href{https://github.com/Muhammad-Osama/robustRisk}{github}.

\section{Numerical experiments}

We illustrate the generality of our framework by addressing four common problems in regression, classification, unsupervised learning and parameter estimation. For the sake of comparison, we also evaluate the recently proposed robust \severalg{} method \cite{diakonikolas19sever}, which was derived on very different grounds as a means of augmenting gradient-based learning algorithms with outlier rejection capabilities. We use the same threshold settings for the \severalg{} algorithm as were used in the experiments in \cite{diakonikolas19sever}, with $\corrub$ in lieu of the unknown fraction $\corr$.

\subsection{Linear Regression}

\begin{figure}[t!]
    \centering
    \begin{subfigure}{0.9\linewidth}
    \includegraphics[width=0.9\linewidth]{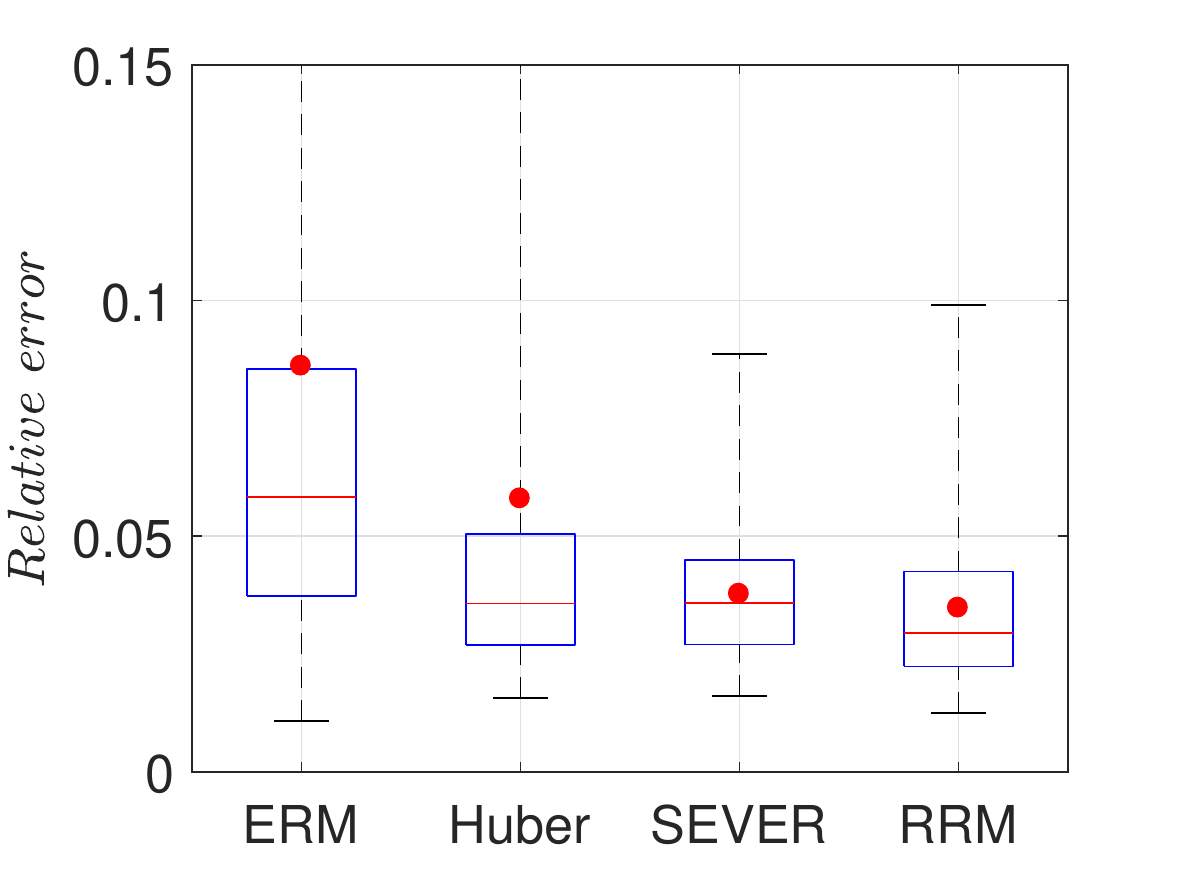}
    \caption{}
    \label{fig:linReg boxplot}
    \end{subfigure}
    \begin{subfigure}{0.9\linewidth}
    \includegraphics[width=0.9\linewidth]{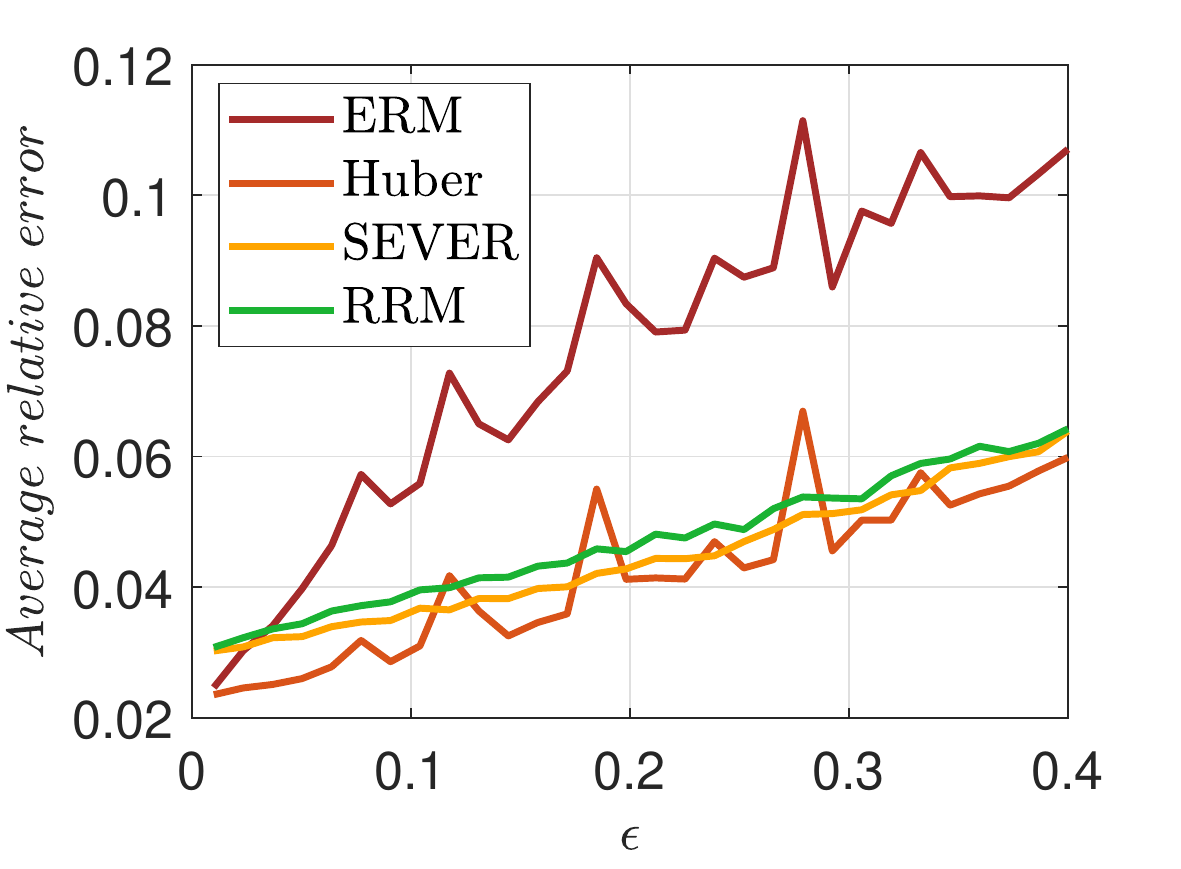}
    \caption{}
    \label{fig:linReg_mse_vs_eps}
    \end{subfigure}
    \caption{Linear regression. a) Box plot of distribution of relative error $\| \bestmodel - \widehat{\parm} \|/\| \bestmodel \|$ when $\corr=20\%$. Each box spans the  $25^{th}$ to $75^{th}$ quantiles and the red dots show the means. The whiskers extend to the minimum and maximum values of error. b) Expected relative error versus percentage of corrupted samples $\corr$. Throughout we use the upper bound $\corrub  = 40 \%$.}
    \label{fig:linReg}
\end{figure}

Consider data $\z~=(\xvec, \y)$, where $\xvec\in\R^{10}$ and $\y\in\R$ denote feature vectors and outcomes, respectively. We consider a class of predictors $\widehat{\y} = \xvec^{\T} \parm$,  where $\parmset = \mathbb{R}^{10}$,  and a squared-error predictive loss $\loss_{\parm}(\xvec,\y) = (\y - \xvec^{\T} \parm )^2$. This loss function targets thin-tailed distributions with a linear conditional mean function.

We learn $\parm$ using $n~=~40$ i.i.d training samples drawn from
\begin{equation}
    \obspdf(\xvec,\y)=(1-\corr)p(\xvec)\truepdf(\y|\xvec)+\corr p(\xvec)\corrpdf(\y|\xvec),
\end{equation}
where 
\begin{equation}\label{eq:linReg dataGen}
    \truepdf(\y|\xvec)=\mathcal{N}(\xvec^{\T}\bestmodel,~\sigma^2),~~\corrpdf(\y|\xvec)=t(\xvec^{\T}\bestmodel, \nu ),
\end{equation}
and $\obspdf(\xvec) = \mathcal{U}([-5,~5]^{10})$. The above data generator yields observations concentrated around a hyperplane, where roughly $\corr$ observations are corrupted by heavy-tailed t-distributed noise.  Data is generated with $\bestmodel~=~\mbf{1}$ and noise standard deviation $\sigma~=~0.25$.

We evaluate the distribution of estimation errors $\| \bestmodel - \widehat{\parm}\|$ relative to $\| \bestmodel \|$ using $100$ Monte Carlo runs. In the first experiment, we set $\corr$ to 20\% and $\nu~=~1.5$, in which case the tails of $q(y| \xvec)$  are so heavy that the variance is undefined. We apply \rrm{} with $\corrub~=~0.40$, which is a conservative upper bound. Note that $\optparm(\widetilde{\prb})$ is a weighted least-squares problem with a closed-form solution. The distribution of errors for $\erm$, \severalg{} and  \rrm{} are  summarized in Figure~\ref{fig:linReg boxplot}. We also include the Huber method, which is tailored specifically for linear regression \citep[ch.~2.6.2]{zoubir2018robust}.  Both \rrm{} and \severalg{} perform similarly in  this  case and are substantially better  than \erm{}, reducing the errors by almost a half. 
 
Next, we study the performance as the percentage of corrupted data $\corr$ increases from $0\%$ to $40\%$. We set $\nu~=~2.5$ so that the variance of the corrupting distribution is defined.  Figure \ref{fig:linReg_mse_vs_eps} shows the expected relative error against $\corr$ for the different methods,  where the robust methods, once again, perform similarly to one another, and much better than \erm{}.

\subsection{Logistic Regression}

\begin{figure*}[t!]
    \centering
    \begin{subfigure}{0.3\linewidth}
    \includegraphics[width=0.9\linewidth]{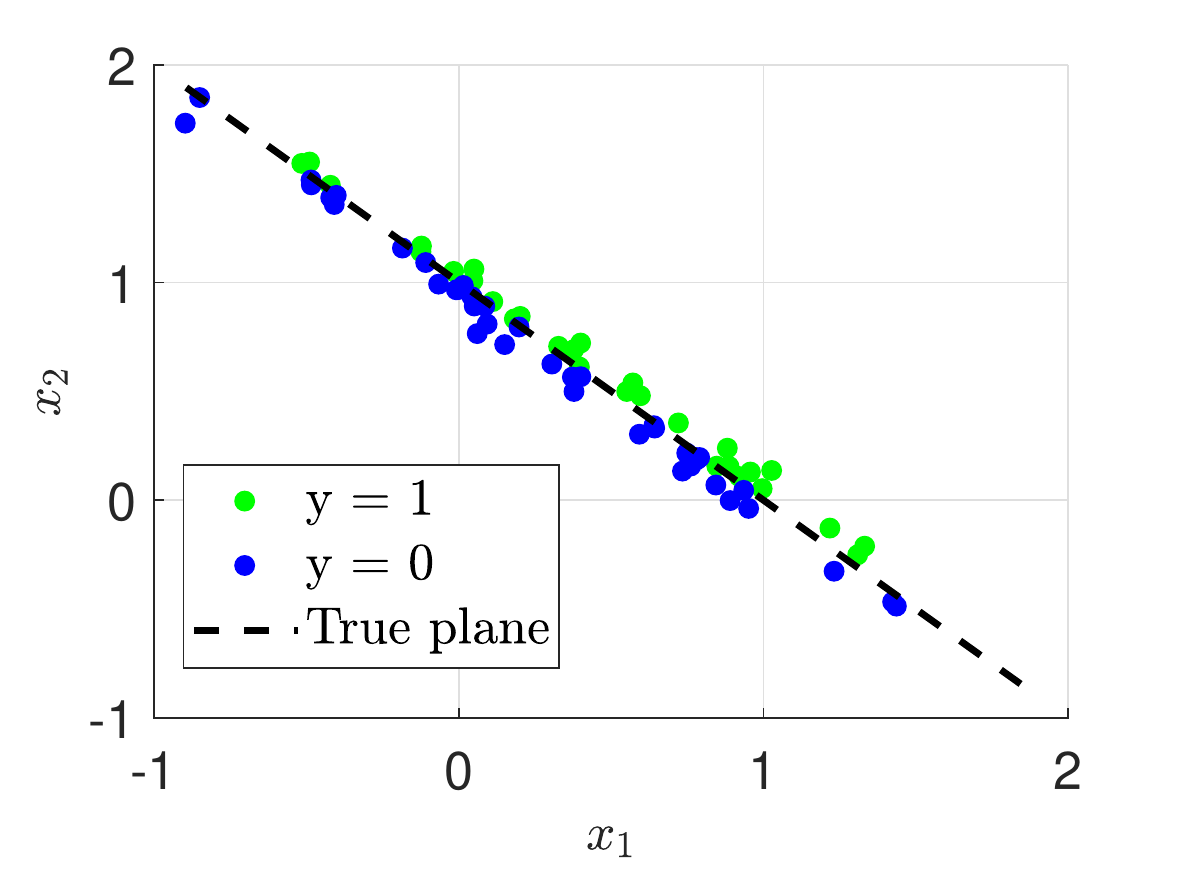}
    \caption{}
    \label{fig:logReg true}
    \end{subfigure}
    \begin{subfigure}{0.3\linewidth}
    \includegraphics[width=0.9\linewidth]{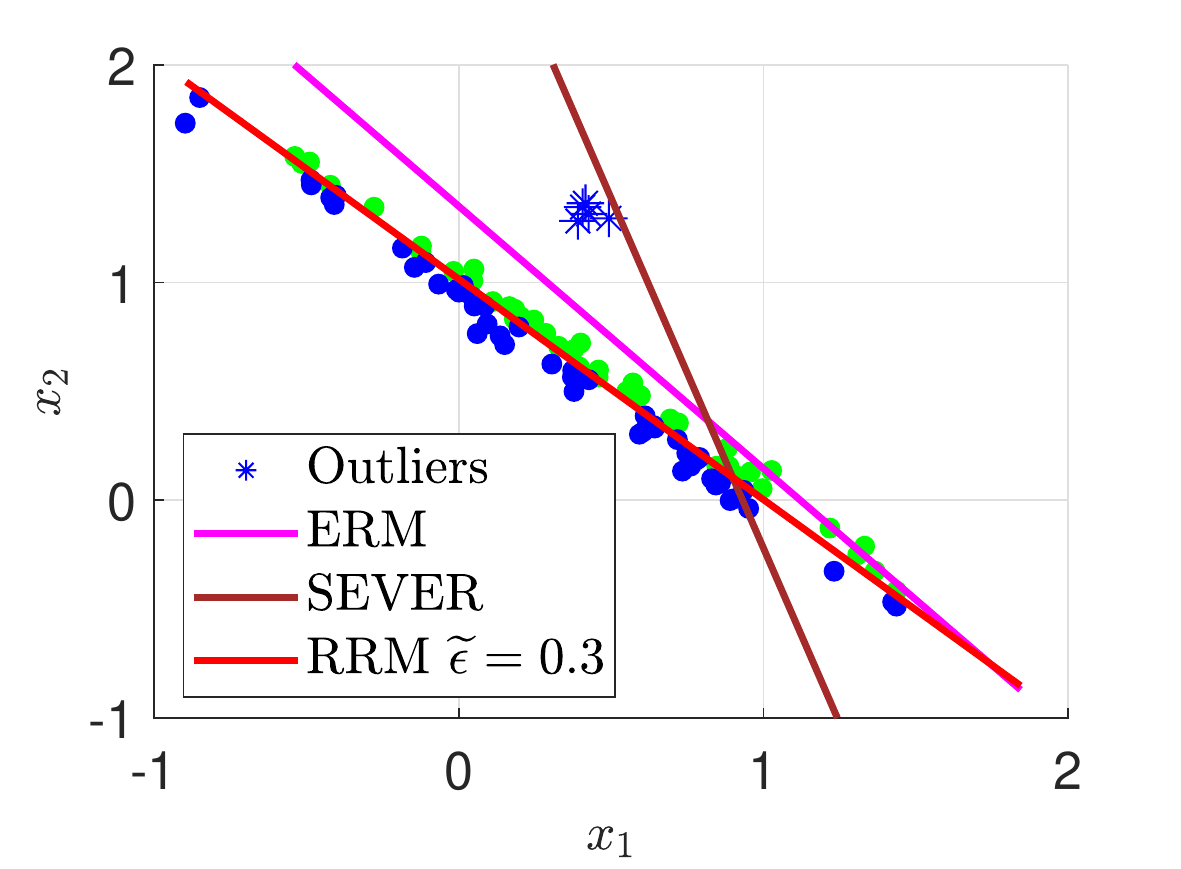}
    \caption{}
    \label{fig: logReg single rlz}
    \end{subfigure}
    \begin{subfigure}{0.3\linewidth}
    \includegraphics[width=0.9\linewidth]{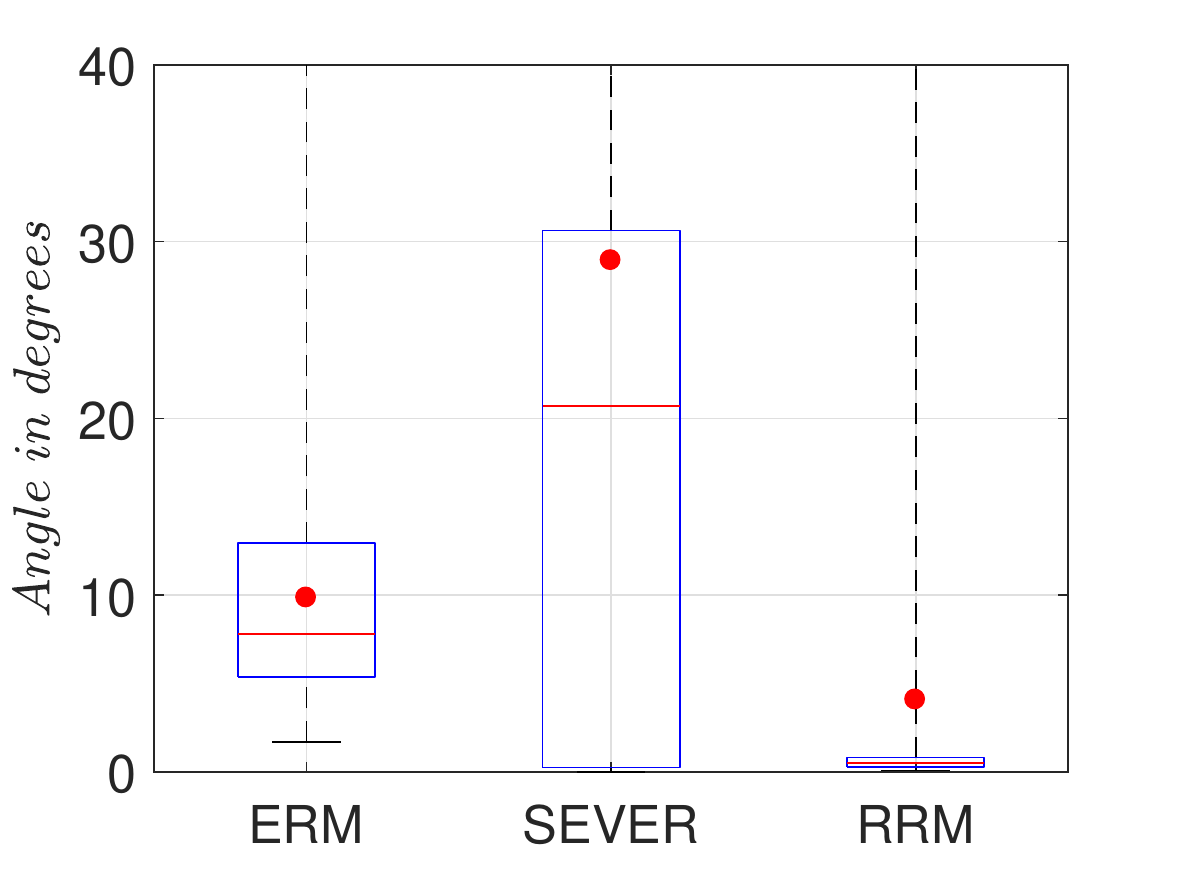}
    \caption{}
    \label{fig:logReg boxplot}
    \end{subfigure}
    \caption{Logistic regression: data points with labels $0$ and $1$ are shown in blue and green, respectively. (a) Single realization from target distribution $\truepdf(\z)$ with linearly separable classes with true hyperplane $\bestmodel$,  (b) Samples from corrupting distribution $\corrpdf(\z)$ (denoted by stars) along with estimated separating hyperplanes $\widehat{\parm}$ using \erm{} and robust \severalg{} and \rrm{} methods. (d) Box plot of angle (in degrees) between the true hyperplane $\bestmodel$ and estimated hyperplanes $\widehat{\parm}$ for different methods.} 
    \label{fig:logReg}
\end{figure*}

Consider data $\z~=~(\xvec,\y)$ where $\xvec\in\R^{2}$ is a feature vector and $\y\in\{0,~1\}$ an associated class label. We consider the cross-entropy loss
\begin{equation} \label{eq:logReg cross entrp loss}
    \loss_{\parm}(\xvec,\y)=-\y\ln\Big(\sigma_{\parm}(\xvec)\Big)-(1-\y)\ln\Big(1-\sigma_{\parm}(\xvec)\Big),
\end{equation}
where $$\sigma_{\parm}(\xvec) = \left(1+ \exp(-\mbs{\phi}^{\T}(\xvec)\parm )\right)^{-1}$$
and $\mbs{\phi}(\xvec)~=~[1,~\xvec]^{\T}$. Thus the loss function targets distributions with linearly separable classes. 

We learn $\parm \in \parmset = \mathbb{R}^3$ using $n~=~100$ i.i.d points drawn from
\begin{equation}\label{eq:dataGen logReg}
    \obspdf(\xvec,\y)~=~(1-\corr)\truepdf(\xvec)\truepdf(\y|\xvec)+\corr\corrpdf(\xvec)\corrpdf(\y|\xvec), 
\end{equation}
where $\truepdf(\xvec)~=~\mathcal{N}\Bigg(\begin{pmatrix}
0.5\\
0.5
\end{pmatrix},\begin{pmatrix}
0.25 & -0.25\rho\\
-0.25\rho & 0.25
\end{pmatrix}\Bigg)$ 
with $\rho~=~0.99$.  An illustration of $\truepdf(\xvec,\y)$ is given in Figure~\ref{fig:logReg true}, where the separating hyperplane corresponds to $\bestmodel~=~[-1,~1,~1]$. The corrupting distribution  is given by $\corrpdf(\xvec)~=~\mathcal{N}\Bigg(\begin{pmatrix}
0.5\\
1.25
\end{pmatrix},\begin{pmatrix}
0.01 & 0\\
0 & 0.01
\end{pmatrix}\Bigg)$ and $\corrpdf(\y~=~0|\xvec)~\equiv~1$ as illustrated in  Figure~\ref{fig: logReg single rlz}.

Data is generated according to \eqref{eq:dataGen logReg} with $\corr$ equal to 5\%.

We apply \rrm{} with $\corrub~=~0.30$. Note that $\optparm(\widetilde{\prb})$ is readily computed using the standard iterative re-weighted least square or MM algorithms \cite{bishop2006pattern}, with minor modifications to take into account the fact that the data points are weighted by $\prb$. Figure \ref{fig: logReg single rlz} shows the learned separating planes, parameterized by $\parm$, for a single realization. We observed that the plane learned by \erm{} and the robust \severalg{} is shifted towards the outliers. By contrast, the proposed \rrm{} method is marginally affected by the corrupting distribution. Figure \ref{fig:logReg boxplot} summarizes the distribution of angles between $\bestmodel$ and $\widehat{\parm}$, i.e., $\arccos{\frac{\widehat{\parm}^{\T}\bestmodel}{||\widehat{\parm}||~||\bestmodel||}}$, using $100$ Monte Carlo simulations.  \rrm{} outperforms the other two methods in this case.

\subsection{Principal Component Analysis}

Consider data $\z\in\R^2$ where we assume $\z$ to have zero mean. Our goal is to approximate $\z$ by projecting it onto a subspace. We consider the loss $\loss_{\parm}(\z)~=~||\z-\mbs{P}_{\parm}\z||^{2}_2$ where $\mbs{P}_{\parm}$ is an orthogonal projection matrix. The loss function targets  distributions where the data is concentrated around a linear subspace. In the case of a one-dimensional subspace $\mbs{P}_{\parm} =\parm\parm^{\T}$, where $\parmset = \{ \parm \in \mathbb{R}^2 : \| \parm \| =1 \}$.

We learn $\parm$ using $n~=~40$ i.i.d datapoints drawn from
\begin{equation}\label{eq:pca dataGen}
    \obspdf(\z)=(1-\corr)\underset{\truepdf(\z)}{\underbrace{\truepdf(z_2|z_1)\truepdf(z_1)}}+\corr\corrpdf(\z),
\end{equation}
where 
\begin{equation}
   \truepdf(z_2|z_1)~=~\mathcal{N}(2z_1,~\sigma^2),~~\truepdf(z_1)~=~\mathcal{N}(0,~1)
\end{equation}
and $\corrpdf(\z)=\tpdf(\mbf{0},\mbf{I},\nu)$ for outliers. Note that $\truepdf(\z)$ in \eqref{eq:pca dataGen} corresponds to a subspace parameterized by $\bestmodel~=~[\frac{1}{\sqrt{5}},~\frac{2}{\sqrt{5}}]^{\T}$.

\begin{figure}[t!]
    \centering
    \includegraphics[width=0.81\linewidth]{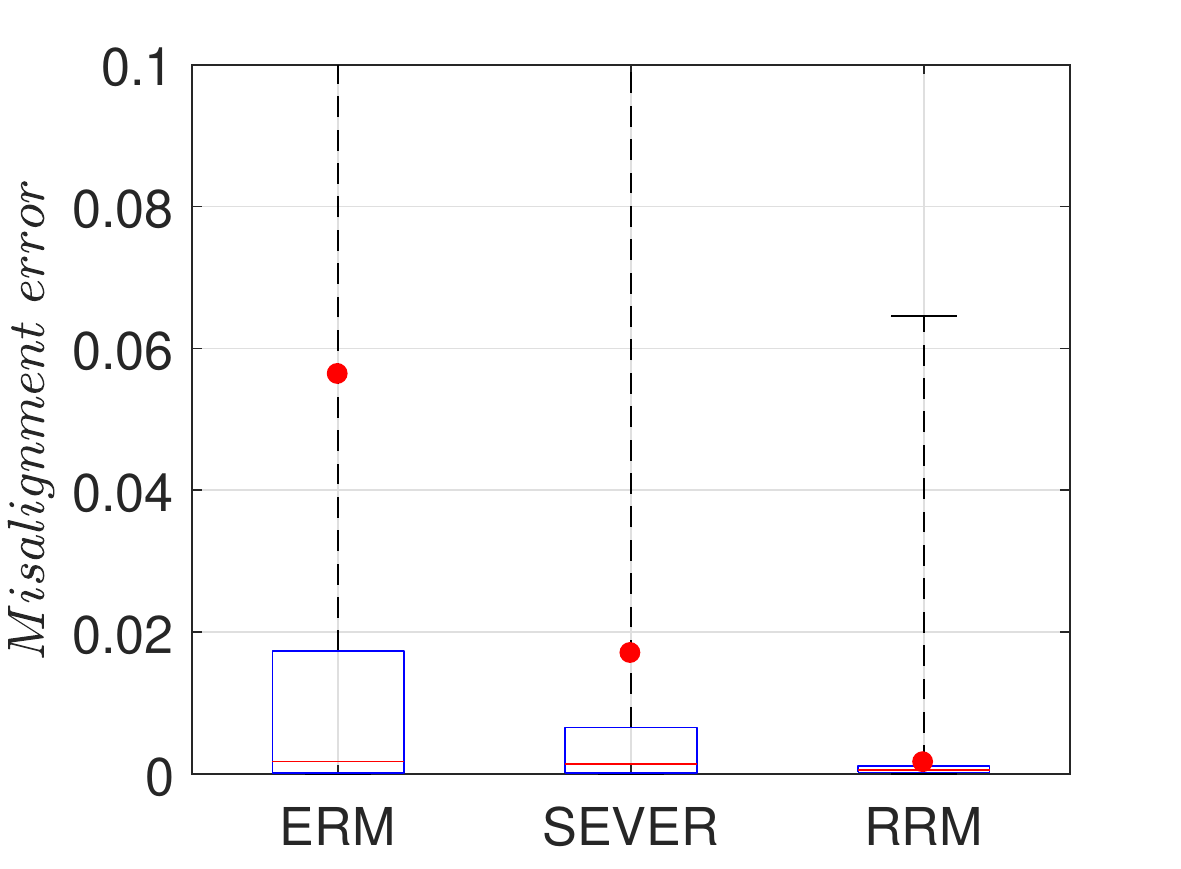}
    \caption{Principal component analysis. Box plot of subspace misalignment error $1-|\cos(\widehat{\parm}^{\T}\bestmodel)|$. }
    \label{fig:pca box plot}
\end{figure}

Data is generated with $\sigma~=~0.25$, $\nu~=~1.5$ and $\corr$ is set to 20\%. We apply \rrm{} with $\corrub~=~0.40$. Note that $\optparm(\widetilde{\prb})$ can be obtained as
\begin{equation}\label{eq:pca eigen prob}
    \optparm(\widetilde{\prb})=\argmax_{\parm \in \parmset}~ \parm^{\T}\mbs{R}\parm,
\end{equation}
which is equivalent to maximizing the Rayleigh quotient and the solution is simply the dominant eigenvector of the covariance matrix
\begin{equation}
    \mbs{R}=\sum_{i=1}^{n}\widetilde{\prbsc}_i\z_{i}\z_{i}^{\T}.
\end{equation}
We evaluate the misalignment of the subspaces using the metric $1-|\cos(\widehat{\parm}^{\T}\bestmodel)|$ evaluated over $100$ Monte Carlo  simulations.  Figure \ref{fig:pca box plot} summarizes the distribution of errors for the three different methods. For this problem, \rrm{} outperforms both \erm{} and \severalg{}.


\subsection{Covariance Estimation}

Consider data $\z\in\R^{2}$ with an unknown mean $\mbs{\mu}$ and covariance $\mbs{\Sigma}$. We consider the loss function 
$$\loss_{\parm}(\z)~=  -(\z-\mbs{\mu})^{\T}\mbf{\Sigma}^{-1}(\z-\mbs{\mu}) + \ln|\mbf{\Sigma}|$$
where $\parm~=~(\mbs{\mu},~\mbf{\Sigma})$. This loss function targets sub-Gaussian distributions.

We learn $\parm$ using $n~=~50$ i.i.d samples drawn from
\begin{equation}\label{eq:dataGen covEst}
    \obspdf(\z)=(1-\corr)\truepdf(\z)+\corr\corrpdf(\z)
\end{equation}
where $\truepdf(\z)~=~\mathcal{N}(\mbs{\mu},\mbs{\Sigma}^{\star})$ and $\corrpdf(\z)~=~\tpdf(\mbs{\mu},\mbs{\Sigma}^{\star},\nu)$.  
Data is generated using \eqref{eq:dataGen covEst} with $\mbs{\mu}~=~\mbf{0}$ and $\mbs{\Sigma}^{\star}~=
\begin{pmatrix}
1&0.8\\
0.8&1\\
\end{pmatrix}
$, and with $\corr~=~20\%$. We set $\nu~=~1.5$, which means that the corrupting distribution $\corrpdf(\z)$ has no finite covariance matrix.

\begin{figure}[t!]
    \centering
    \includegraphics[width=0.81\linewidth]{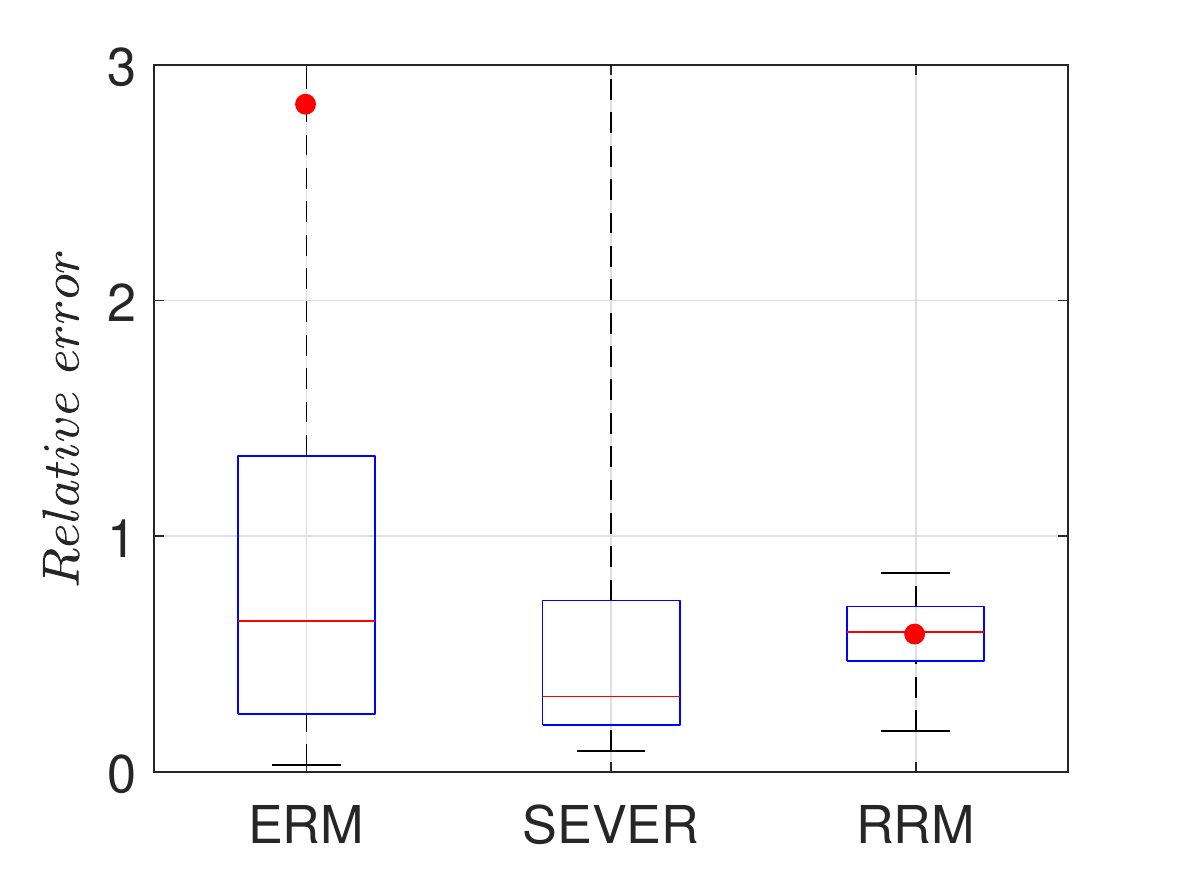}
    \caption{Covariance estimation. Box plot of distribution of relative errors $\| \mbs{\Sigma}^*  -\mbs{\Sigma} \|_F/\| \mbs{\Sigma}^*\|_F$. Note that the expected relative error for \severalg{} is too large to be contained in the given plot.}
    \label{fig:covEst boxplot}
\end{figure}

We apply \rrm{} with upper bound $\corrub~=~0.30$. Note that $\optparm(\widetilde{\prb})$ has a closed-form solution, given by the weighted sample mean and covariance matrix with the weight vector equal to $\widehat{\prb}$. We evaluate the error $||\mbf{\Sigma}^{\star}-\widehat{\mbf{\Sigma}}||_{F}$ relative to $||\mbf{\Sigma}^{\star}||_{F}$ over $100$ Monte Carlo simulations and show it in Figure \ref{fig:covEst boxplot}. We see that \severalg{} is prone to break down due to the heavy-tailed outliers, whereas \rrm{} is stable.

\section{Real data}

Finally, we test the performance of \rrm{} on real data. We use the Wisconsin breast cancer dataset from the UCI repository \cite{cancerData}. The dataset consists of $n~=~683$ points, with features $\xvec\in\R^{9}$ and labels $\y\in\{0,~1\}$. The class labels $0$ and $1$ correspond to `benign' and `malignant' cancers, respectively. $60\%$ of the data was used for training, which was subsequently corrupted by flipping the labels of $40$ class $1$ datapoints to $0$ ($\corr\approx 10\%$). The goal is to estimate a linear separating plane to predict the class labels of test data. We use the cross-entropy loss function $\loss_{\parm}(\z)$ in \eqref{eq:logReg cross entrp loss} and apply the proposed $\rrm$ method with $\corrub= 0.15$. For comparison, we also use the standard $\erm$ and the robust \severalg{} methods.

Tables \ref{Table:real erm} for \erm, \ref{Table:real sever} for \severalg{} and \ref{Table:real rrm} for \rrm{} summarize the results using the confusion matrix as the metric. The classification accuracy for the $\rrm$ method is visibly higher than that of $\erm$ and \severalg{} for class $1$.

\begin{table}[h!]
\centering
\begin{tabular}{ | c | c | c | } 
\hline
$n~=~274$ & Predicted $1$ & Predicted $0$ \\ 
\hline
Actual $1$ & 69 & 28 \\ 
\hline
Actual $0$ & 1 & 176 \\ 
\hline
\end{tabular}
\caption{Confusion matrix for $\erm$. Classification accuracy $89.42\%$.}
\label{Table:real erm}
\end{table}

\begin{table}[h!]
\centering
\begin{tabular}{ | c | c | c | } 
\hline
$n~=~274$ & Predicted $1$ & Predicted $0$ \\ 
\hline
Actual $1$ & 71 & 26 \\ 
\hline
Actual $0$ & 3 & 174 \\ 
\hline
\end{tabular}
\caption{Confusion matrix for \severalg{}. Classification accuracy $89.42\%$.}
\label{Table:real sever}
\end{table}

\begin{table}[h!]
\centering
\begin{tabular}{ | c | c | c | } 
\hline
$n~=~274$ & Predicted $1$ & Predicted $0$ \\ 
\hline
Actual $1$ & 76 & 21 \\ 
\hline
Actual $0$ & 3 & 174 \\ 
\hline
\end{tabular}
\caption{Confusion matrix for $\rrm$. Classification accuracy $91.24\%$.}
\label{Table:real rrm}
\end{table}

\section{Conclusion}
We proposed a general risk minimization approach which provides robustness in a wide range of statistical learning problems in cases where a fraction of the observed data comes from a corrupting distribution. Unlike existing general robust methods, our approach does not depend on any problem-specific thresholding techniques to remove the corrupted data points, as are used in existing literature, nor does it rely on a correctly specified corruption fraction $\corr$. We illustrated the wide applicability and  performance of our method by testing it on several classical supervised and unsupervised statistical learning problems using both simulated and real data. 







\nocite{langley00}

\bibliography{mybib}
\bibliographystyle{icml2020}


\end{document}